\documentclass[
]{ceurart}

\usepackage{listings}
\usepackage{todonotes}
\usepackage{threeparttable}
\usepackage{subcaption}
\usepackage{longtable}
\usepackage[capitalize]{cleveref}
\usepackage{caption}
\usepackage{array}
\DeclareCaptionFormat{customlongtable}{#1\\#3}

\makeatletter
\newif\if@anonymize
\@anonymizefalse  

\newcommand{\anonymize}[2]{%
  \if@anonymize
    #1  
  \else
    #2  
  \fi
}
\makeatother

\begin{document}

\copyrightyear{2024}
\copyrightclause{Copyright for this paper by its authors.
  Use permitted under Creative Commons License Attribution 4.0
  International (CC BY 4.0).}

\conference{CHR24: 5th Conference on Computational Humanities Research,
  December 04--06, 2024, Aarhus, Denmark}

\title{Visual Navigation of Digital Libraries: Retrieval and Classification of Images in the National Library of Norway’s Digitised Book Collection}


\author[1]{Marie Roald}[%
orcid=0000-0002-9571-8829,
email=marie.roald@nb.no,
]
\cormark[1]
\address[1]{Research and Special Collections, The National Library of Norway, Norway}

\author[1]{Magnus Breder Birkenes}[
email=magnus.birkenes@nb.no,
]
\author[1]{Lars Gunnarsønn Bagøien Johnsen}[
email=lars.johnsen@nb.no,
]
\cortext[1]{Corresponding author.}

\begin{abstract}
Digital tools for text analysis have long been essential for the searchability and accessibility of digitised library collections. Recent computer vision advances have introduced similar capabilities for visual materials, with deep learning-based embeddings showing promise for analysing visual heritage. Given that many books feature visuals in addition to text, taking advantage of these breakthroughs is critical to making library collections open and accessible. In this work, we present a proof-of-concept image search application for exploring images in the National Library of Norway’s pre-1900 books, comparing Vision Transformer (ViT), Contrastive Language-Image Pre-training (CLIP), and Sigmoid loss for Language-Image Pre-training (SigLIP) embeddings for image retrieval and classification. Our results show that the application performs well for exact image retrieval, with SigLIP embeddings slightly outperforming CLIP and ViT in both retrieval and classification tasks. Additionally, SigLIP-based image classification can aid in cleaning image datasets from a digitisation pipeline.
\end{abstract}


\maketitle

\section{Introduction}
With the goal of preserving and disseminating Norwegian cultural heritage, the National Library of Norway (NLN) began digitising its collection in 2006. This collection, acquired per the Norwegian Legal Deposit Act\footnote{\url{https://lovdata.no/dokument/NL/lov/1989-06-09-32}}, spans various materials, including books, newspapers, journals, posters, radio, movies and more \cite{birkenesNBDHLABCorpus2023}. Almost all books and most newspapers have already been digitised, barring a few exceptions, and the current focus is on processing newspapers, journals, and non-text-based media \cite{birkenesNBDHLABCorpus2023}. However, digitisation alone is insufficient to make cultural heritage available; it is also necessary to ensure that the digitised content is easy to view and access is not overly restricted. Thus, the \emph{Bokhylla} agreement grants regulated access \cite{kopinorBokhyllaavtalenFra20242023}, and the online library \emph{Nettbiblioteket} lets users view collections with an International Image Interoperability Framework (IIIF) \cite{snydmanInternationalImageInteroperability2015} based viewer and perform full-text searches using Elasticsearch. Finally, NLN offers limited access to the textual content through NB DH-LAB \cite{birkenesNBDHLABCorpus2023} and corresponding webapps\footnote{\url{https://www.nb.no/dh-lab/apper/}} which provides tools based on text aggregates (e.g. n-grams, collocations and concordances) to facilitate automated and reproducible analysis of the text. 

Currently, these tools have largely been based on text extracted from Analysed Layout and Text Object-Extensible Markup Language (ALTO-XML) files\footnote{\url{https://www.loc.gov/standards/alto/}} generated by optical character recognition (OCR) models during digitisation \cite{birkenesDigitalLibraryNgrams2015}. However, the output XML also contains coordinates for graphical elements. These graphical elements represent non-textual elements in the books, e.g. illustrations or decorations. While such elements are an important part of the books, they have been cumbersome to explore, requiring manual inspection. Therefore, an essential missing step for making NLN’s digitised collection more accessible is making these graphical elements easier to explore and analyse.

An approach to make such elements explorable, is creating tools for image search, either in the form of exact image retrieval (i.e. recovering a specific image) or semantic image retrieval (i.e. recovering an image with similar contents) or both. While text-based search engines are commonplace, image search is more complicated \cite{meiMultimediaSearchReranking2014a,zhouRecentAdvanceContentbased2017}. Early methods matched images using surrounding text \cite{meiMultimediaSearchReranking2014a}, but this approach demands high-quality textual descriptions, which can be lacking. Alternatively, exact image retrieval traditionally relies on handcrafted image features for comparison \cite{meiMultimediaSearchReranking2014a,zhouRecentAdvanceContentbased2017}. Handcrafting such features can be challenging, and typically form a dense vector, which can hinder efficient lookups.

However, recent technological advancements have simplified the implementation of image search engines. Various tools now implement efficient search indices for dense vectors, such as the hierarchical navigable small worlds (HNSW) index \cite{malkovEfficientRobustApproximate2020}. Moreover convolutional neural networks (CNNs) and vision transformers (ViTs) have alleviated the need for handcrafted image features for computer vision \cite{heDeepResidualLearning2016,dosovitskiyImageWorth16x162021}. Furthermore, there has been an influx of multimodal models, like Contrastive Language-Image Pre-training (CLIP) \cite{radfordLearningTransferableVisual2021} and Sigmoid Loss for Language Image Pre-Training (SigLIP) \cite{zhaiSigmoidLossLanguage2023}. The recent advances in computer vision and proliferation of advanced pre-trained computer vision models has empowered the development of new research and tools for exploring and analysing image-based data in the digital humanities \cite{arnoldDistantViewingAnalyzing2019,weversVisualDigitalTurn2019,smitsMultimodalComputationalHumanities2021,hosseiniMapReaderComputerVision2022,smitsMultimodalTurnDigital2023,ruthClustersGraphsScalable2023}.

Previous work on machine learning-driven computer vision-based image search tools for digital humanities mainly focuses on cleanly digitised materials such as collections of videos, photographs, lantern slides and medieval illuminations \cite{arnoldDistantViewingAnalyzing2019,smitsMultimodalComputationalHumanities2021,smitsMultimodalTurnDigital2023,meineckeMedievalDistantViewing2024}. However, there is limited work applying such tools to images extracted from the output of automatic layout detection of scanned media, e.g. books and newspapers. Such image collections pose unique challenges. First, the magnitude of data is often larger than for collections of photographs. Second, such data can contain artefacts not found in cleanly digitised materials. For example, detected bounding boxes might be inaccurate. False positives can occur, where the automatic layout detection mistakenly marks, e.g. tables or blank pages, as graphical elements. Avoiding such artefacts can be infeasible, as redoing layout analysis for a collection of sizeable magnitude can be cost-prohibitive and not guaranteed to succeed. Therefore, a natural next step is exploring machine learning-based image retrieval in the context of NLN’s collection of scanned automatically processed media.

This short paper details ongoing work on these challenges, with three primary contributions:
\begin{enumerate}
    \item Developing a proof-of-concept image search application for NLN’s pre-1900 books.
    \item Comparing modern image embeddings for image retrieval in NLN’s digitised books.
    \item Evaluating pre-trained models for fine-tuned classification of image categories.
\end{enumerate}

\section{Background and related work}
Two traditional approaches for image retrieval are context-based full-text search --- querying the images’ textual context --- and hashing-based approaches for exact image retrieval. The former typically works by using an inverted index to efficiently retrieve relevant images via e.g. term frequency-inverse document frequency (TF-IDF) weighting \cite{sparckjonesStatisticalInterpretationTerm1972}, before potentially re-ranking them based on image features \cite{meiMultimediaSearchReranking2014a}. The hashing-based alternative works by computing a compact hash, or “fingerprint”, that can be used for efficient exact image retrieval \cite{biswasStateArtImage2021}. 

More recent image retrieval approaches compute image similarities using deep learning-based image classification models such as ViTs \cite{dosovitskiyImageWorth16x162021} or CNNs \cite{heDeepResidualLearning2016}. These models first transform an image into an embedding, which is used as input for a logistic regression model. The key insight in using these models for image retrieval is that we can compute image similarities by comparing the embeddings, e.g. with the cosine similarity.

However, by using classification models, we assume that embeddings learned by training on image-label combinations are informative enough to group images semantically, which can hinder generalisation to out-of-sample images \cite{mayoHowHardAre2023}. Another approach is multimodal models like CLIP and SigLIP. In short, these models work by combining an image transformer and a text transformer to compute image and text embeddings -- aligning them to ensure strong cosine similarity for matching pairs. This approach has been successfully applied to e.g. image retrieval and zero-shot classification \cite{radfordLearningTransferableVisual2021}, and generalise better to out-of-sample images \cite{radfordLearningTransferableVisual2021,mayoHowHardAre2023}.

During CLIP and SigLIP training, models receive shuffled image-caption pairs and compute probabilities for matches. Such training demands extensive data and computational resources. To circumvent this, it is common to use pre-trained models and the popularity of model repositories, like Huggingface Hub \cite{wolfTransformersStateoftheArtNatural2020} and Torch Hub \cite{anselPyTorchFasterMachine2024}, has made using models trained on massive datasets accessible.

While methods for efficient sparse vector queries have existed for decades \cite{knuthArtComputerProgramming1997}, querying based on image embeddings requires dense vector queries, which is still a research topic. However, the recently proposed HNSW-index for approximate nearest neighbour search \cite{malkovEfficientRobustApproximate2020} has gained traction for accuracy and efficiency. The index consists of a hierarchy of navigable small world graphs \cite{malkovApproximateNearestNeighbor2014}, each built from different data subsets, and querying consists of iteratively traversing the hierarchy, enabling efficient navigation through large datasets.

Applying modern computer vision to problems in digital humanities has recently gained traction. The term \emph{distant viewing} is introduced in \cite{arnoldDistantViewingAnalyzing2019}, which demonstrates how computer vision methods for clustering and object detection can be applied to image- and video-data. Building on this, \cite{weversVisualDigitalTurn2019} shows how CNN-based semantic image retrieval can be used to explore trends in newspaper advertisements and illustrations extracted from Delpher — a digitised materials search engine by the Dutch national library. Moreover, \cite{meineckeMedievalDistantViewing2024} demonstrate how a combination of monomodal image- and language-models can be used to combine and enrich two manually annotated collections of medieval illuminations and \cite{smitsMultimodalComputationalHumanities2021,smitsMultimodalTurnDigital2023} shows how a CLIP model can be used to explore and label magic lantern slides efficiently and that it can struggle with zero-shot classification of old illustrations. Using CLIP embeddings, \cite{ruthClustersGraphsScalable2023} clusters news videos and employs a graph-based approach for efficient exploration. Machine learning-driven image retrieval tools for libraries and museums, like Maken\footnote{\url{https://www.nb.no/maken/}} , Bildsök\footnote{\url{https://lab.kb.se/bildsok/}} and Nasjonalmuseet Beta\footnote{\url{https://beta.nasjonalmuseet.no/collection/}} have also emerged. These previous works highlight computer vision's potential in digital humanities, and thus, evaluating and comparing such models in the context of NLN’s digitised book collection is a relevant next step.

\section{Methods}
\subsection{Extracting images}
To search the images, they must first be extracted from the digitised book collection. During NLN’s digitisation, books are scanned and processed through a pipeline including layout detection and OCR, producing ALTO-XML files\footnote{\url{https://digitalpreservation-blog.nb.no/docs/formats/preferred-formats-en/}} named after Uniform Resource Names (URNs). These files contain page information, describing the page in terms of four block types: \lstinline{TextBlock}, \lstinline{Illustration}, \lstinline{GraphicalElement} and \lstinline{CompositeBlock} (blocks containing other blocks)\footnote{\url{https://www.loc.gov/standards/alto/techcenter/layout.html}}. In the ALTO-XML files parsed for this work, all illustrations and graphical elements are tagged as GraphicalElement. Parsing these files, we extracted the page URN, coordinates, and size for each graphical element in addition to the textual context of each image in the digitised books. For this work, we processed pre-1900 books, creating a sufficiently large, yet manageable subset for testing.

For each graphical element, we used NLN’s IIIF API\footnote{\url{https://iiif.io/api/image/2.0/}} to download images from URLs following the format in \cref{tab:iiif.url}, discarding images with aspect-ratio \(\geq 50\). By integrating ALTO-XML files with the IIIF endpoint — both technologies already utilised by NLN — we obtained images from digitised Norwegian books before 1900.

\begin{table}[]
    \centering
    \caption{The IIIF URL format.} \label{tab:iiif.url}
    \begin{tabular}{@{}>{\raggedleft\arraybackslash}p{0.37\linewidth-0.37em}@{\hspace{1em}}p{0.63\linewidth-0.63em}@{}}
        \toprule
        Description & Example \\
        \midrule
        Scheme & \texttt{https://} \\
        Prefix & \texttt{www.nb.no/services/image/resolver/} \\
        Identifier (\texttt{URN}) & \texttt{URN:NBN:no-nb\_digibok\_2009070210001\_0618/} \\
        Region (\texttt{left,top,width,height}) & \texttt{430,432,2195,2348/} \\
        Size (\texttt{width,height}) & \texttt{274,294/} \\
        Rotation (\texttt{degrees}) & \texttt{0/}\\
        Filename (\texttt{filename.filetype}) & \texttt{default.jpg}\\
        Full URL & \url{https://www.nb.no/services/image/resolver/URN:NBN:no-nb\_digibok\_2009070210001\_0618/430,432,2195,2348/274,294/0/default.jpg}\\
        \bottomrule
    \end{tabular}
\end{table}

\subsection{Creating the vector search application}
We computed image embeddings using Huggingface Transfomers \cite{wolfTransformersStateoftheArtNatural2020} with three models: ViT (\lstinline{google/vit-base-patch16-224}\footnote{Commit hash: \lstinline{3f49326eb077187dfe1c2a2bb15fbd74e6ab91e3}}), CLIP (\lstinline{openai/clip-vit-base-patch32}\footnote{Commit hash: \lstinline{3d74acf9a28c67741b2f4f2ea7635f0aaf6f0268}}) and SigLIP (\lstinline{google/siglip-base-patch16-256-multilingual}\footnote{Commit hash: \lstinline{a66c5982c8c396206b96060e2bf837d6731a326f}}). Each pre-trained model’s preprocessing pipeline involved resizing images to the input shapes (224 for ViT and CLIP, and 256 for SigLIP) and scaling the pixel values. For ViT and SigLIP, images were resized to 224 × 224 and 256 × 256 pixels, altering the aspect ratio. CLIP resized the smallest dimension to 224, preserving the aspect ratio, then center-cropped to 224 × 224 pixels. Next, we used the corresponding image transformer and obtained embeddings of sizes 768 (ViT and SigLIP) and 512 (CLIP).

After computing embeddings, we ingested them into a Qdrant database and used FastAPI to create an application programming interface (API) for efficient querying by images, embedding vectors, image IDs, or context-based text search. Qdrant supports fast K-nearest neighbour search for both dense and sparse vectors. For image-based queries, we used a cosine similarity-based HNSW index, and for context-based full-text queries, we used a dot-product-based inverted index for TF-IDF (details in supplement on GitHub\footnote{\url{\anonymize{https://github.com/anonymised/repo-name}{https://github.com/Sprakbanken/CHR24-image-retrieval}}}).  We used default parameters for all search indices. The vector database and the API are hosted on-premise, exposing only the API to the Internet. The application also includes a frontend, implemented using Flask and HTMX, hosted using Google Cloud Run with 512 MiB RAM and one vCPU.

\subsection{Classifying based on embedding vectors}
As the graphical elements stem from NLN's digitisation process, many segmentation anomalies are also tagged as graphical elements. Common examples are blank pages, parts of tables, and text. To estimate the fraction of such regions, we used HumanSignal Label Studio and manually labelled a dataset containing 2000 images as either \emph{Blank page}, \emph{Segmentation anomaly}, \emph{Illustration or photograph}, \emph{Musical notation}, \emph{Map}, \emph{Mathematical chart} or \emph{Graphical element} (e.g. initial, decorative border, etc.).

After labelling the data, we fitted regularised logistic regression models (using scikit-learn v1.5.0 \cite{pedregosaScikitlearnMachineLearning2011}) to classify images based on their embedding vectors. This can be interpreted as a form of transfer learning, fine-tuning the last layer of the transformer model. The embedding vector type (i.e. ViT, CLIP or SigLIP) and the complexity parameter (inverse ridge parameter) were selected using nested cross-validation with 20 outer folds and ten inner folds. Models were selected based on a micro-averaged F1-score (the harmonic mean of micro-averaged precision and sensitivity). We selected the complexity parameter from ten logarithmically spaced values between \(10^{-4}\) and \(10^4\). Finally, we computed the confusion matrix in the outer cross-validation loop (the evaluation loop). The supplement describes the overall cross-validation algorithm in Algorithms 1 and 2.

\subsection{Evaluating searches}
To evaluate the search, we first manually inspected some example queries before performing a systematic evaluation on exact image retrieval. To simulate exact image retrieval scenarios, we selected the 684 images labelled as \emph{Illustration or photograph}, \emph{Map} or \emph{Mathematical chart} as target images, and applied random cropping (\(\leq 15\,\%\), independently on all sides), rotation (\(\pm 0-10\,^\circ\)) and scaling (\(\pm 0-20\,\%\), independently for width and height). Then, querying the database with these transformed images, we evaluated the Top \(N\) accuracy measuring whether our application retrieved the target image in the first result (Top 1), first row (Top 5), first two rows (Top 10) or results at all (Top 50). 

\section{Results}
\begin{figure}[b]
    \centering
    \begin{subfigure}{0.49\textwidth}
        \includegraphics[trim= 40 40 40 80,clip,width=\textwidth]{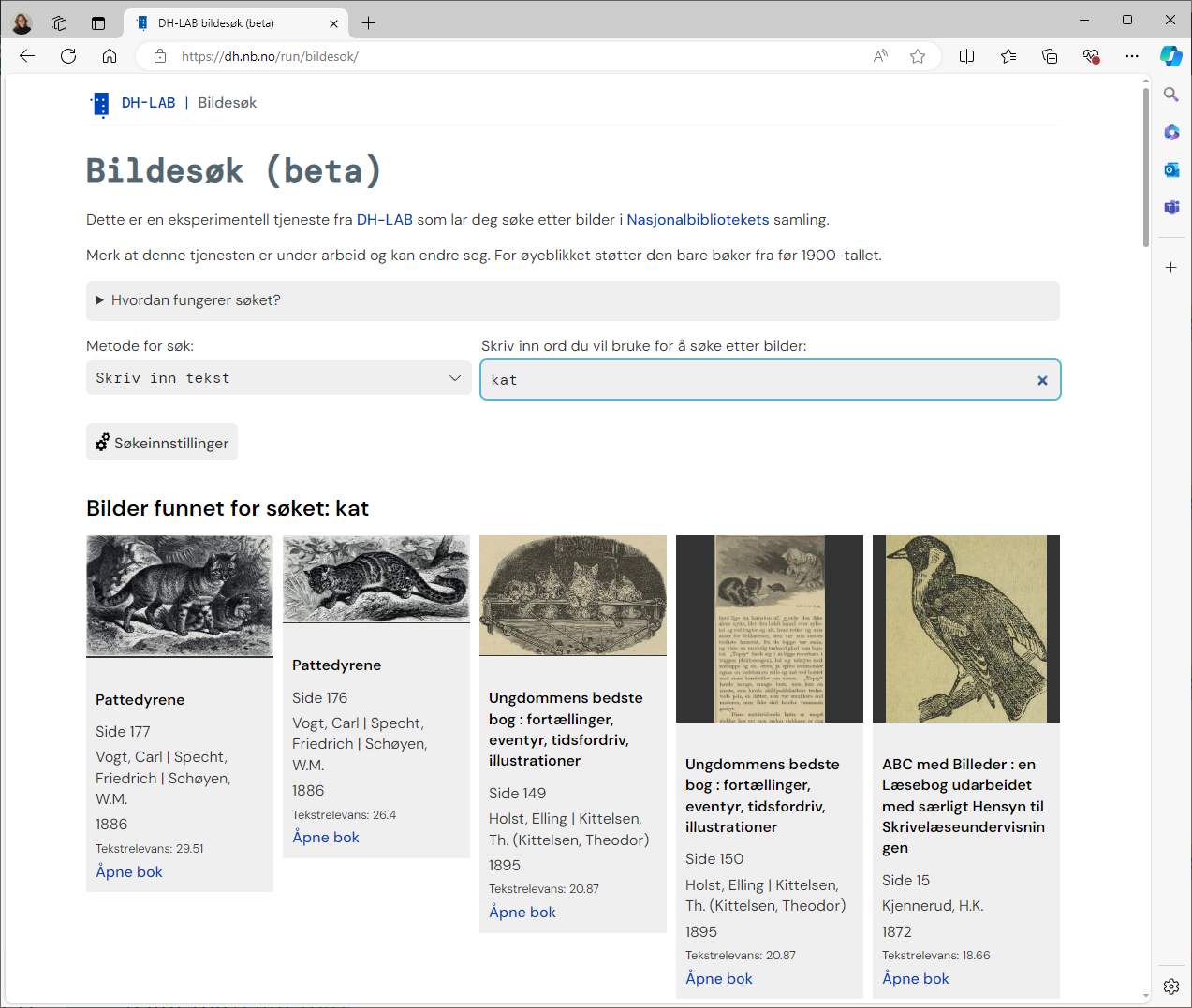}
        \caption{} \label{fig:image.search.a}
    \end{subfigure}
    \begin{subfigure}{0.49\textwidth}
        \includegraphics[trim= 40 40 40 80,clip,width=\textwidth]{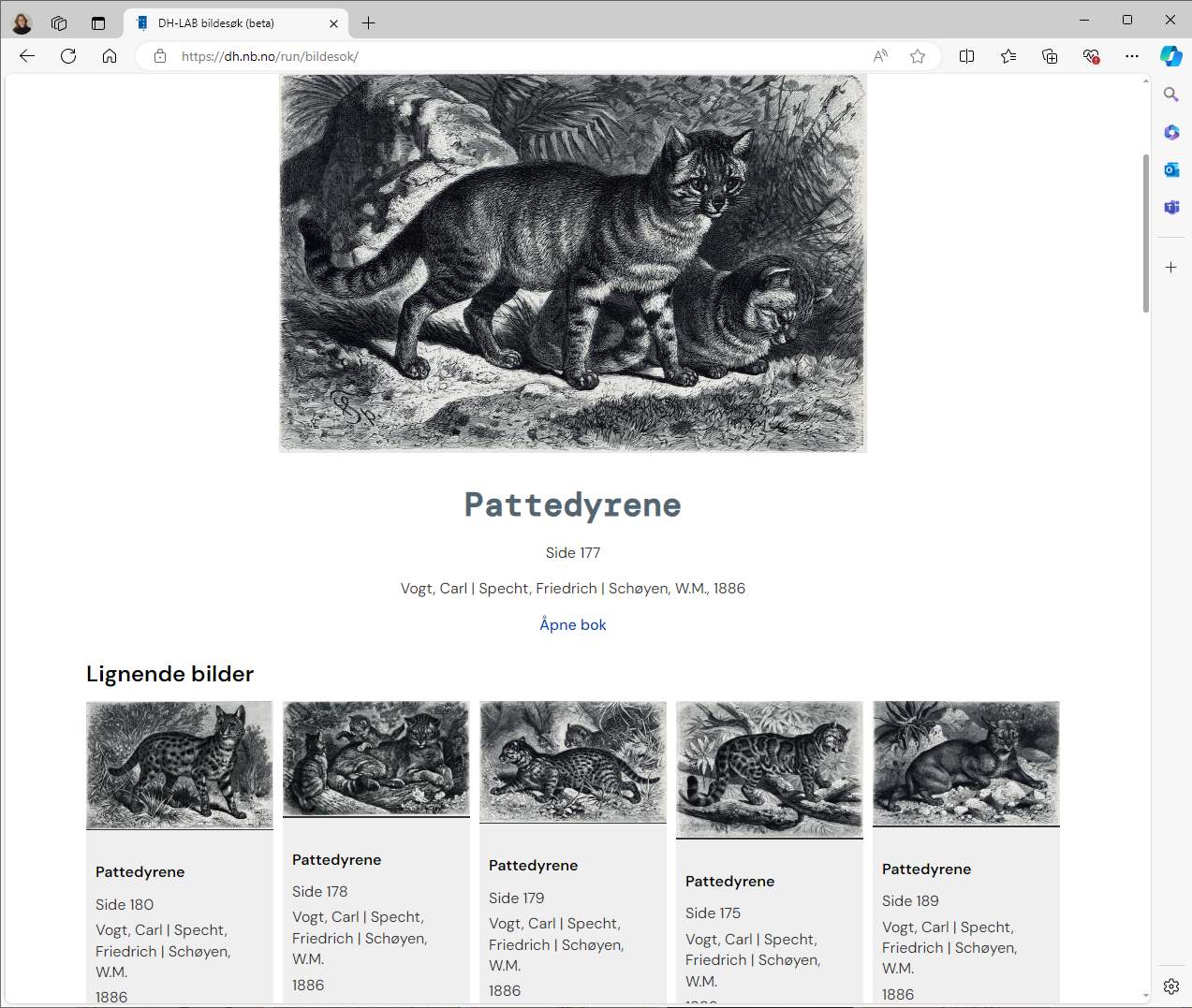}
        \caption{} \label{fig:image.search.b}
    \end{subfigure}
    \begin{subfigure}{0.49\textwidth}
        \includegraphics[trim= 40 40 40 80,clip,width=\textwidth]{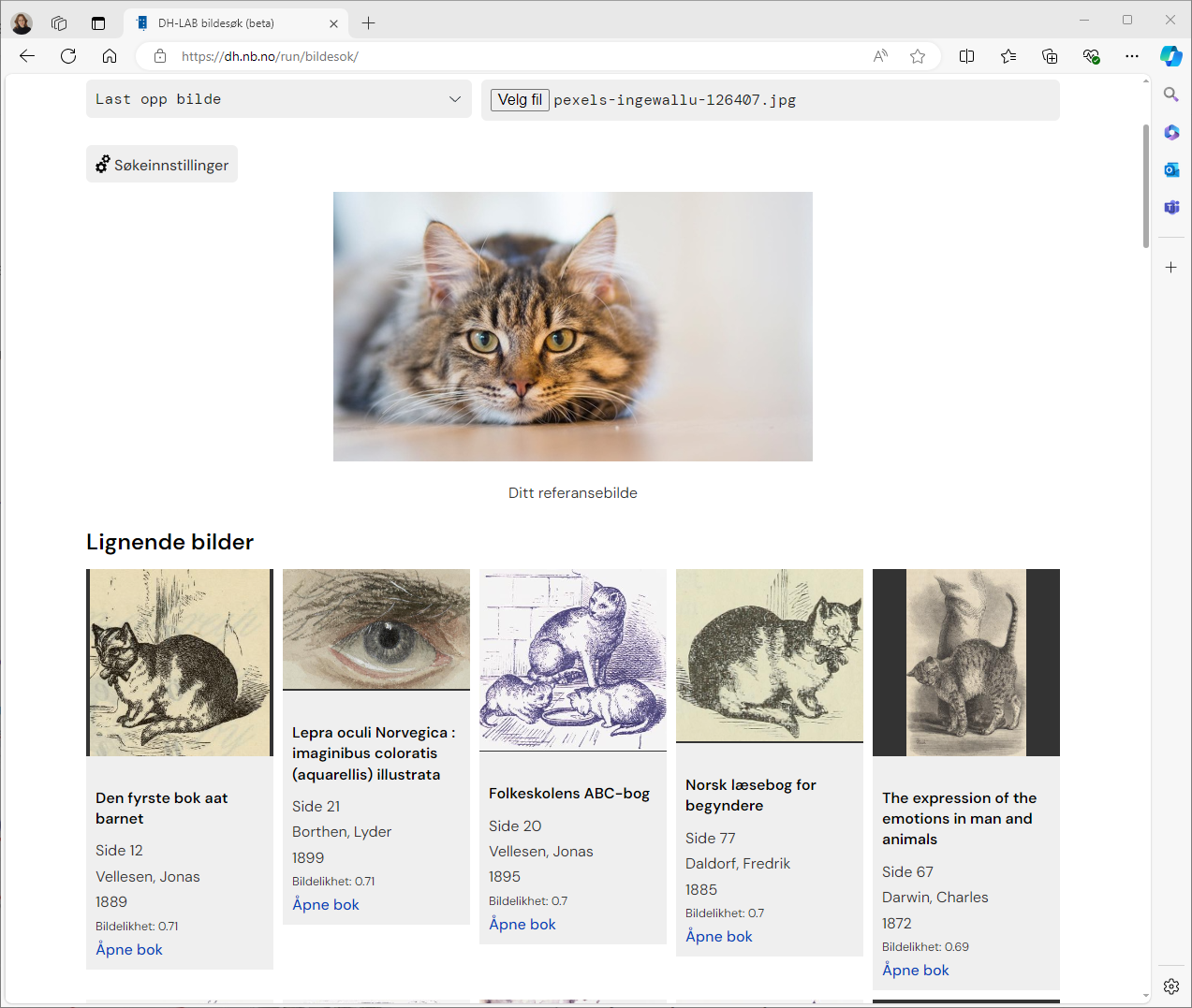}
        \caption{} \label{fig:image.search.c}
    \end{subfigure}
    \begin{subfigure}{0.49\textwidth}
        \includegraphics[trim= 40 40 40 80,clip,width=\textwidth]{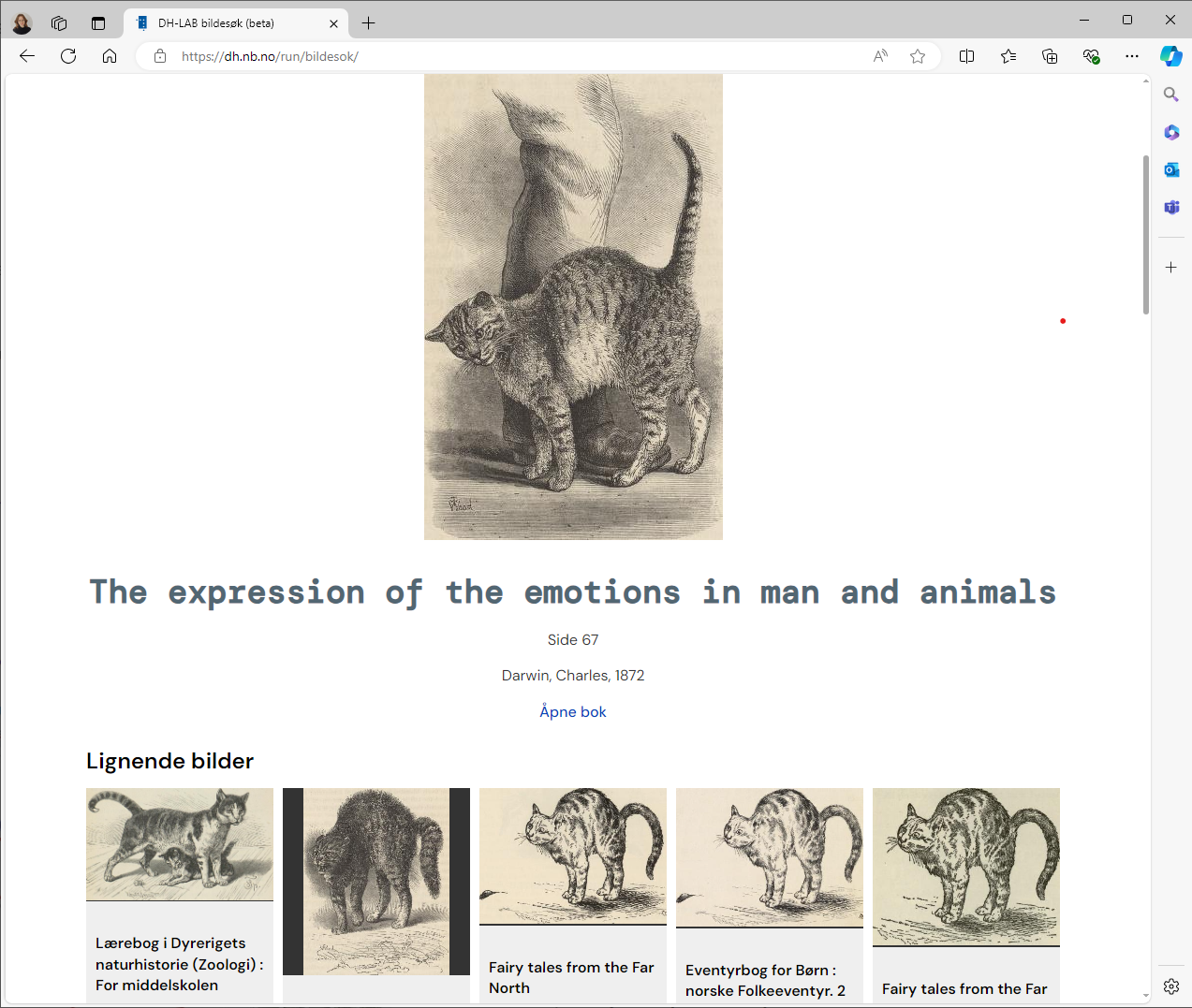}
        \caption{} \label{fig:image.search.d}
    \end{subfigure}
    \caption{Screenshots of the image search application: context-based search for "kat" (old Norwegian for cat) (a) and image-based query with a user-uploaded cat image (c). (b) and (d) show the results when selecting an image in (a) and (c), respectively. The app also has a collapsible sidebar (not shown) that we used for selecting SigLIP embedding vectors.}
    \label{fig:image.search}
\end{figure}

\Cref{fig:image.search} shows screenshots from the application\footnote{\url{\anonymize{https://anonymissed.url/}{https://dh.nb.no/run/bildesok/}}} for image searches using full-text (\cref{fig:image.search.a}) or image similarity (\cref{fig:image.search.b,fig:image.search.c,fig:image.search.d}). \Cref{tab:query.examples} shows image-based query results with four different images. For the first row, the query exists in the collection, and all models recover it as the top result. Similarly, for the second row, all models return nautical results, and CLIP is the only model that does not return illustrations with lighthouses. Finally, the third and fourth rows show examples of querying with images outside of the collection, where we see that the returned images are content-wise similar. The fourth row demonstrates an example where CLIP embedding vectors fail, leading to irrelevant results. Furthermore, the exact image retrieval experiments demonstrate that our application can recover queried transformed images. As demonstrated in \cref{tab:reverse.image.search}, SigLIP performed slightly better than ViT and CLIP and retrieved \(94\,\%\) of the target images in the first two rows of the search and  \(97\,\%\) in all ten displayed rows. See GitHub for code and details.

\begin{figure}[t]
\begin{subfigure}[b]{0.34\textwidth}
\begin{tabular}{@{}lrr@{}}
\toprule
Dataset & Train & Full\(^\star\) \\
Label &  &  \\
\midrule
\makecell[lt]{Map} & 44 & 7692 \\
\makecell[lt]{Mathematical\\chart} & 48 & 10184 \\
\makecell[lt]{Musical\\notation} & 113 & 25398 \\
\makecell[lt]{Graphical\\element} & 330 & 72858 \\
\makecell[lt]{Blank\\page} & 349 & 66336 \\
\makecell[lt]{Segmentation\\anomaly} & 524 & 110254 \\
\makecell[lt]{Illustration or\\photograph} & 592 & 129867 \\
\midrule
\makecell[lt]{In total} & 2000 & 422589 \\
\bottomrule
\end{tabular}
\caption{}
\end{subfigure} 
\hfill
\begin{subfigure}[b]{0.64\textwidth}
\includegraphics[trim= 125 0 100 0,clip, width=1\linewidth]{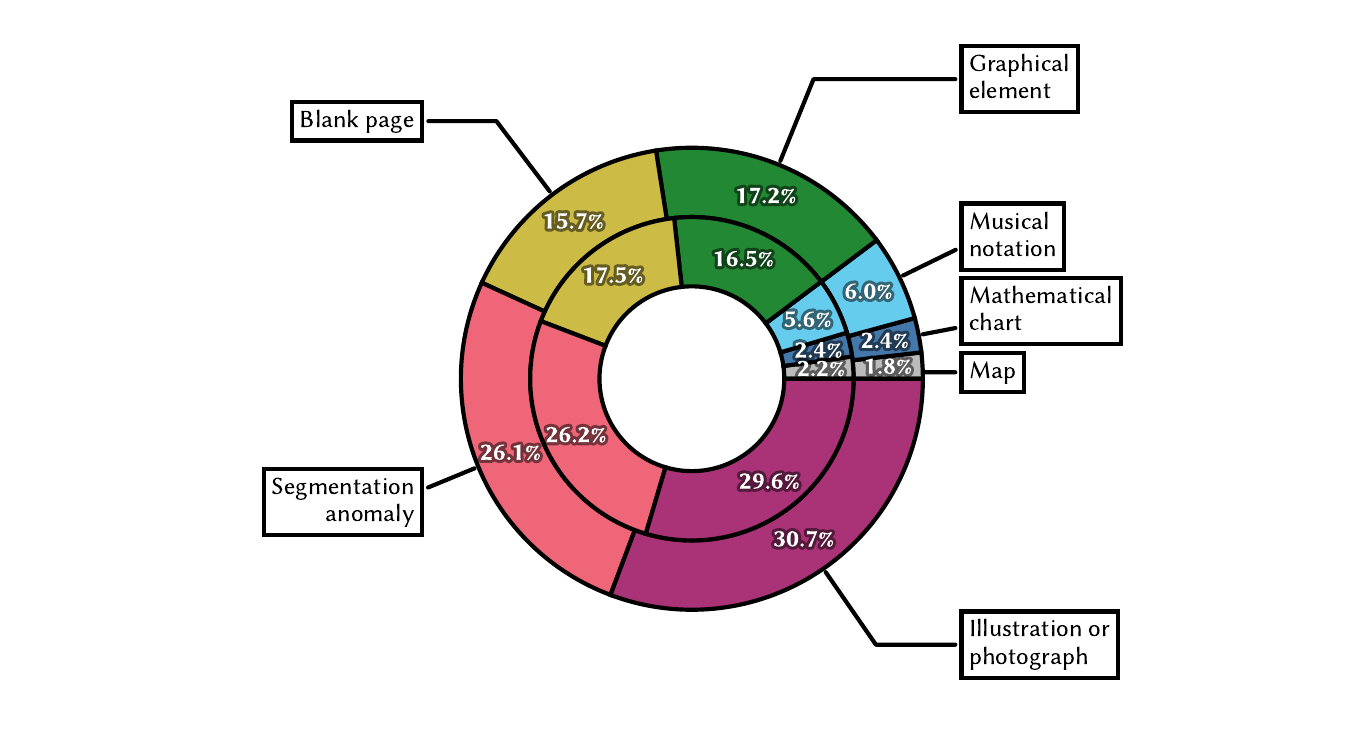}
\caption{}
\end{subfigure}
\caption{The class distribution for the manually labelled training set and estimated class distribution for the full dataset. (a) shows absolute counts, and (b) shows label distributions for the training set (inner) and estimated distributions for the full dataset (outer).}
\label{fig:class.distribution}
\end{figure}

The manual image labelling\footnote{The labels and analysis code are available on GitHub}  showed that 349/2000 (\(17\,\%\)) of the graphical elements were blank pages and 524/2000 (\(26\,\%\)) were segmentation anomalies (e.g. tables, text, etc.) --- for complete label distribution, see \cref{fig:class.distribution}. Moreover, the logistic regression model performs well, obtaining a cross-validated F1 score of \(96\,\%\) (\(\sigma=5.1\,\%\)). From the cross-validated confusion matrix, we see that only 66/1127 (\(<6\,\%\)) of all graphical elements were incorrectly classified as either blank pages or segmentation errors, with a marked amount of incorrect classifications being from the “Graphical element” class. We also observed that the SigLIP embeddings were selected in all 20 outer cross-validation folds, indicating their superiority for this classification task compared to ViT and CLIP. \cref{fig:class.distribution}  also shows the estimated class distribution on the full dataset.

\section{Discussion and conclusion}
These promising results demonstrate that pre-trained computer vision models provide meaningful embeddings. This is notable as our data consists of pre-1900 book images and differs vastly from the training set of such models, which are typically scraped from the internet. Furthermore, the results indicate that SigLIP embeddings slightly outperforms CLIP and ViT for all tasks — even for image classification, which ViT was trained for — in line with prior results showing that multimodal models are more robust to out-of-sample data \cite{mayoHowHardAre2023}.

While all models perform well for retrieval, CLIP sometimes struggled, particularly if the object of interest was off-centre. In such cases, the object is cropped out during preprocessing and matches will be based on the remaining image. Furthermore, the application performs well for exact image retrieval, even with up to \(30\,\%\) cropping in both directions and up to \(\pm 10\,^\circ\) rotation. These results are promising, but more work is still needed to evaluate performance for other degradations (e.g. simulated print and scanning artefacts). Finally, the encouraging image classification results indicate advantages of adding this methodology to the data ingestion pipeline. Filtering out irrelevant elements can save up to \(40\,\%\) storage and improve the search results.

In conclusion, we found that by combining tagged graphical elements of the book digitisation process, NLN’s IIIF endpoint and recent advances in artificial intelligence, we can create an efficient image search application that facilitates exploring the library’s collection in a new way.

\section{Future work}
As the current prototype image-search app only supports books pre-1900, a natural extension is including illustration objects from all NLN's digitised books and newspapers. Moreover, as one use case we consider is exact image retrieval, an obvious next step is more thorough analysis of the the application's accuracy on this task, e.g. using additional evaluation measurements for recall, and including domain-specific degradation (e.g. simulated halftone and scanning artefacts). Another avenue for future work is comparing deep learning-based similarity measures with simpler, less computation- and storage-intensive approaches like hashing-based methods. Additionally, we want to make the software more adaptable, ultimately creating open-source infrastructure to further these methods' accessibility for other ALTO-XML and IIIF collections.

Future work should explore the embeddings further, e.g. using CLIP and SigLIP for text-based image retrieval. Additionally, performance could improve by fine-tuning the embeddings on domain-relevant data. Moreover, we have so far only used the embeddings for image retrieval and classification. Using the embeddings as the base to discover clusters, automatically tag the images or create image descriptions are, therefore, interesting potential steps. Another important direction is digging deeper into what the models consider "similar" through visualisations and empirical experiments. Finally, because deep learning-based embeddings are trained on datasets with known biases \cite{birhaneMultimodalDatasetsMisogyny2021,smitsMultimodalTurnDigital2023,mandalMultimodalBiasAssessing2023}, examining biases in these embeddings is crucial.


\bibliography{main}

\newpage~  
{ \sffamily
\begin{longtable}{@{}lllllll@{}}
\caption{Example of search results using the different models} \label{tab:query.examples}\\    
\toprule
 &  & Pos. 1 & Pos. 2 & Pos. 3 & Pos. 4 & Pos. 5 \\
Query image & Model &  &  &  &  &  \\
\midrule
\endfirsthead
\toprule
 &  & Pos. 1 & Pos. 2 & Pos. 3 & Pos. 4 & Pos. 5 \\
Query image & Model &  &  &  &  &  \\
\midrule
\endhead
\midrule
\multicolumn{7}{r}{Continued on next page} \\
\midrule
\endfoot
\bottomrule
\endlastfoot
\multirow[t]{3}{*}{\includegraphics[width=0.12\textwidth]{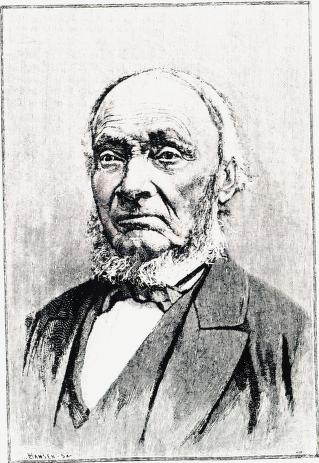}} & SigLIP & \includegraphics[width=0.12\textwidth]{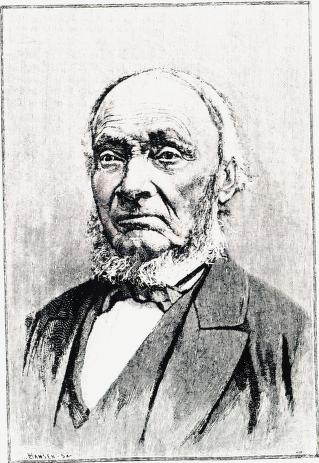} & \includegraphics[width=0.12\textwidth]{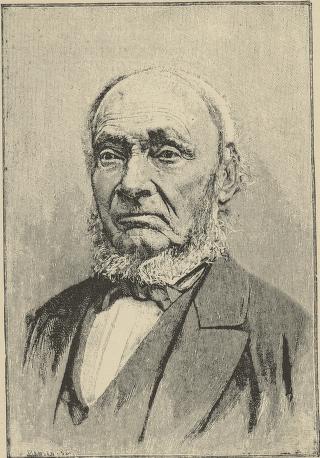} & \includegraphics[width=0.12\textwidth]{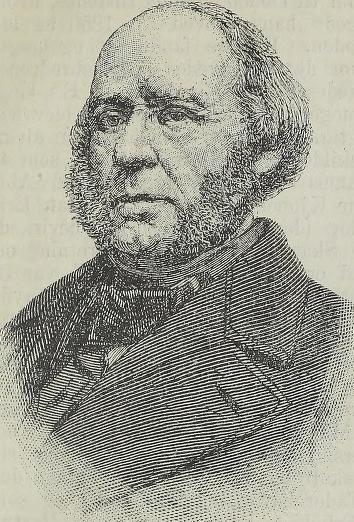} & \includegraphics[width=0.12\textwidth]{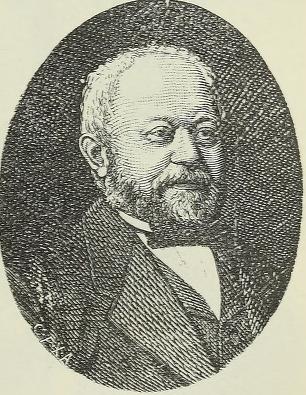} & \includegraphics[width=0.12\textwidth]{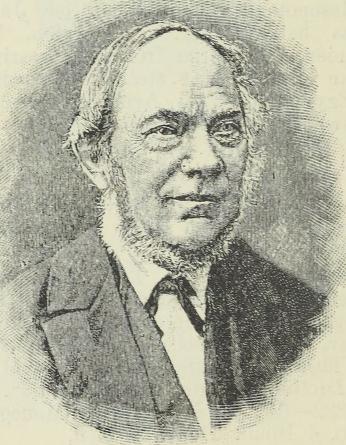} \\*
 & CLIP & \includegraphics[width=0.12\textwidth]{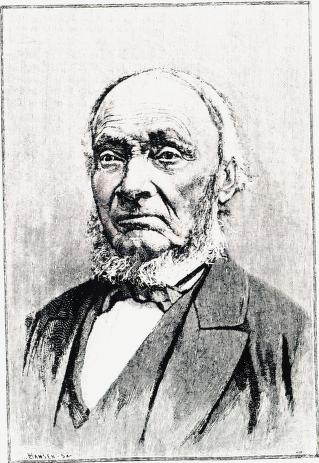} & \includegraphics[width=0.12\textwidth]{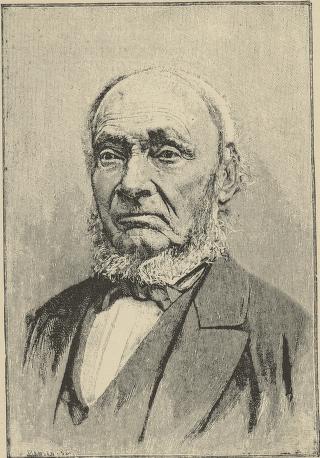} & \includegraphics[width=0.12\textwidth]{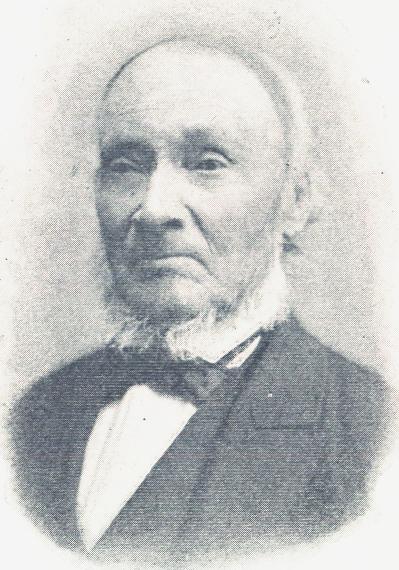} & \includegraphics[width=0.12\textwidth]{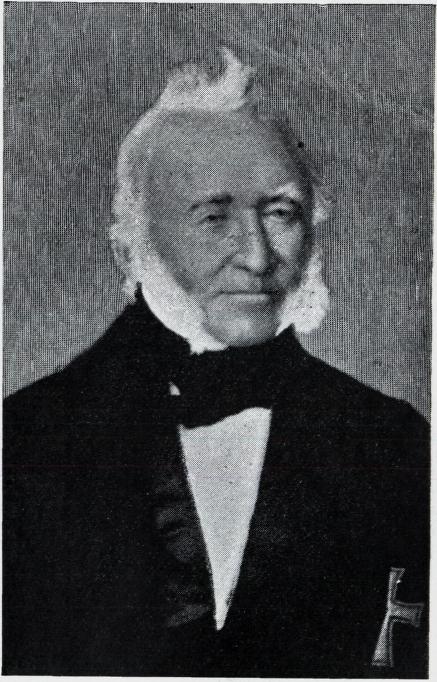} & \includegraphics[width=0.12\textwidth]{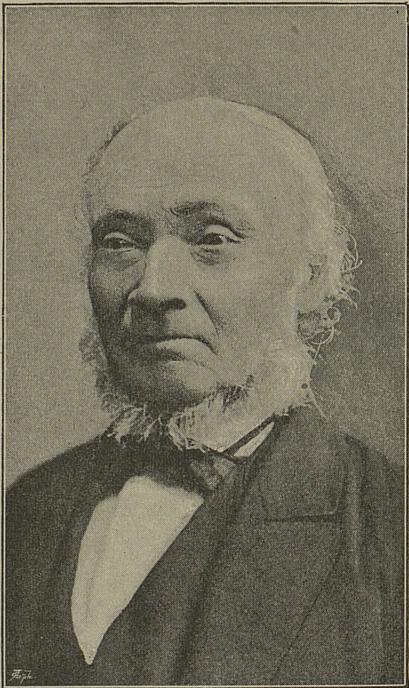} \\*
 & ViT & \includegraphics[width=0.12\textwidth]{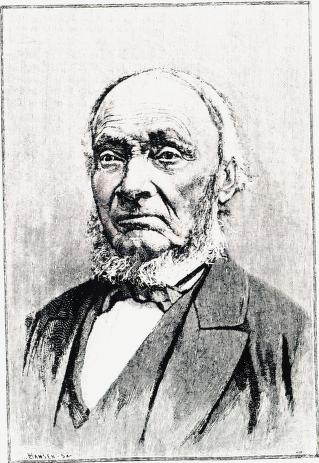} & \includegraphics[width=0.12\textwidth]{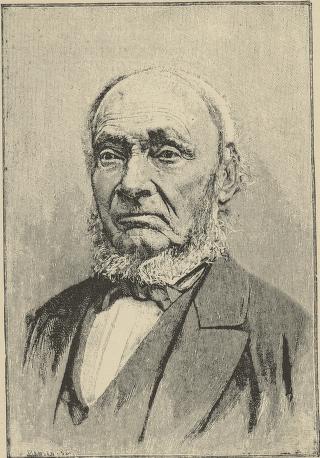} & \includegraphics[width=0.12\textwidth]{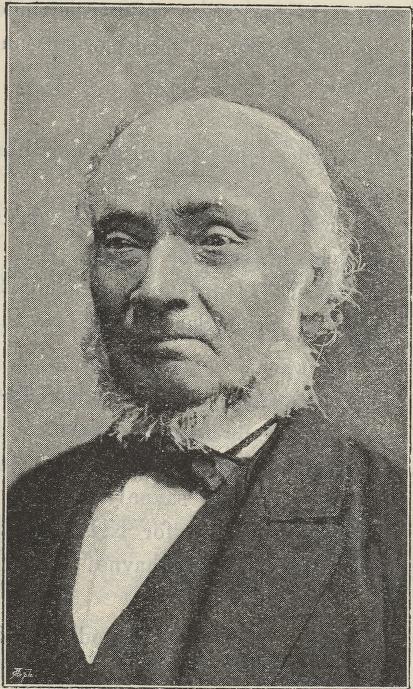} & \includegraphics[width=0.12\textwidth]{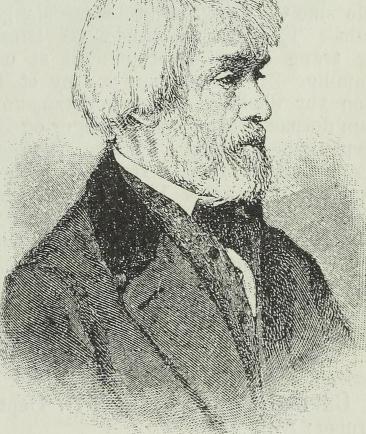} & \includegraphics[width=0.12\textwidth]{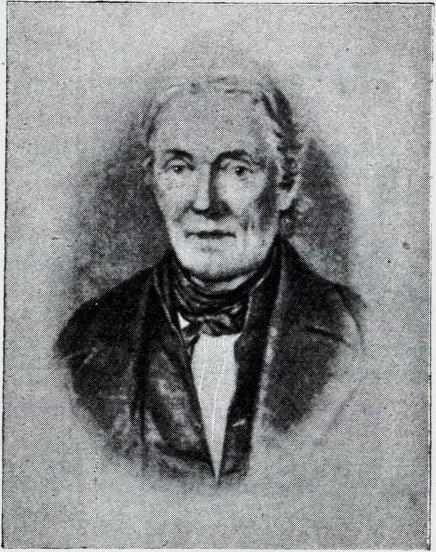} \\*
\hline  \\[-2ex]
\multirow[t]{3}{*}{\includegraphics[width=0.12\textwidth]{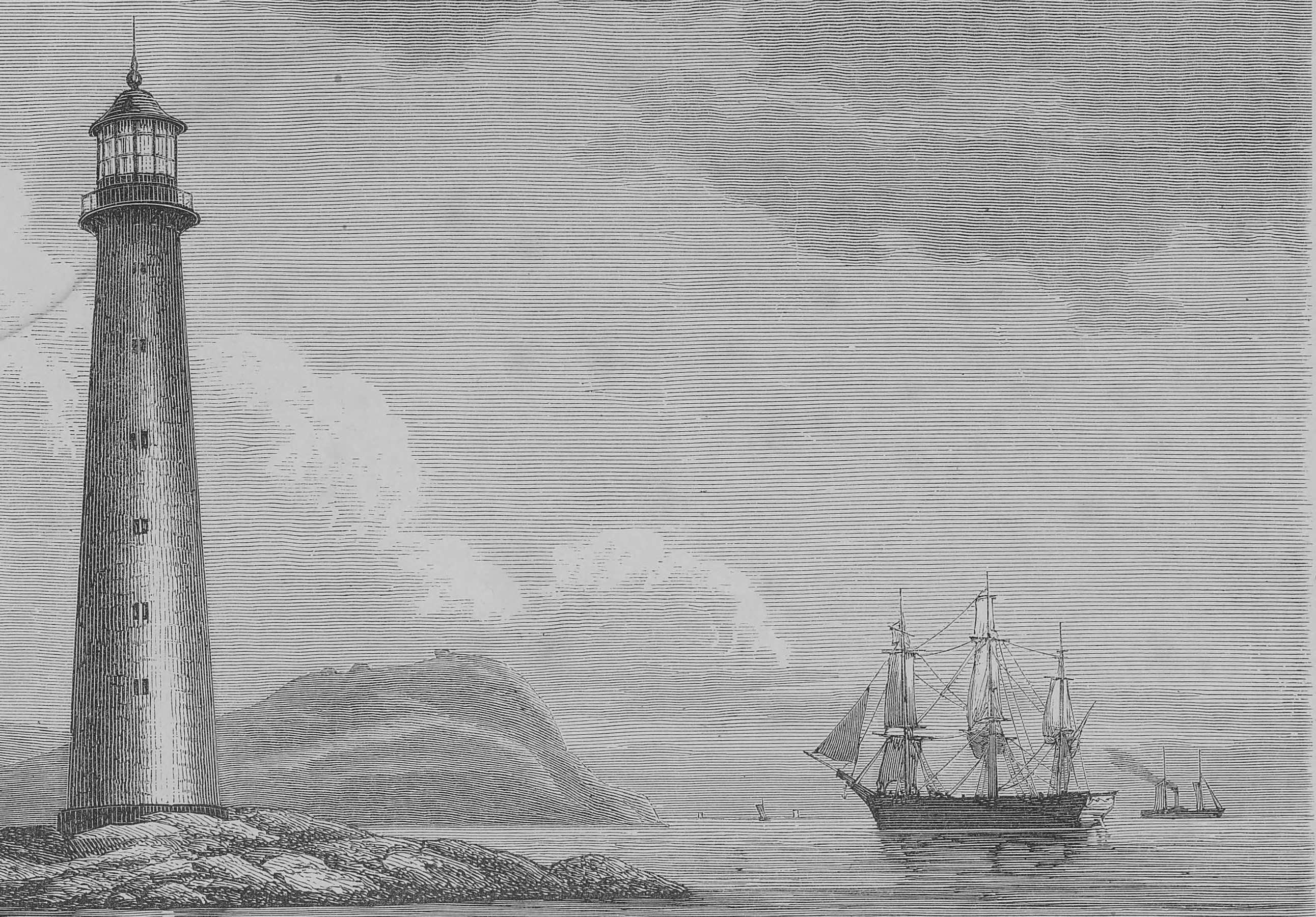}} & SigLIP & \includegraphics[width=0.12\textwidth]{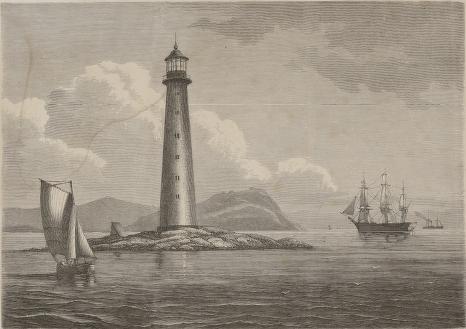} & \includegraphics[width=0.12\textwidth]{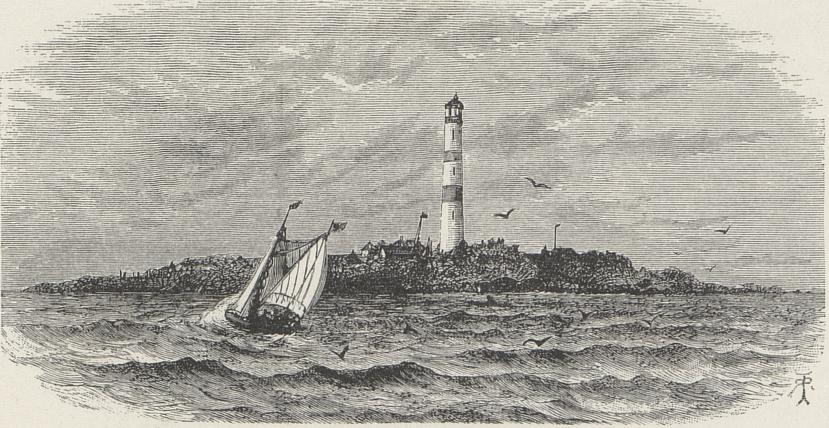} & \includegraphics[width=0.12\textwidth]{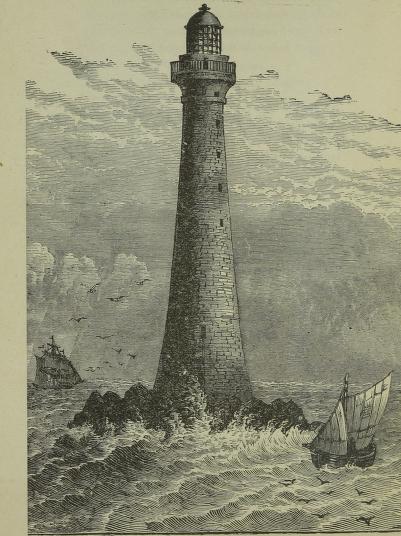} & \includegraphics[width=0.12\textwidth]{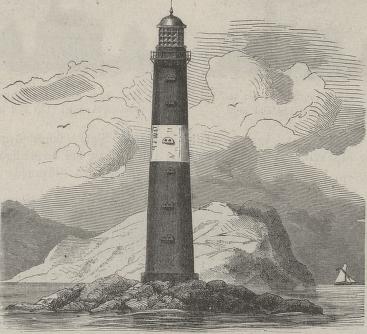} & \includegraphics[width=0.12\textwidth]{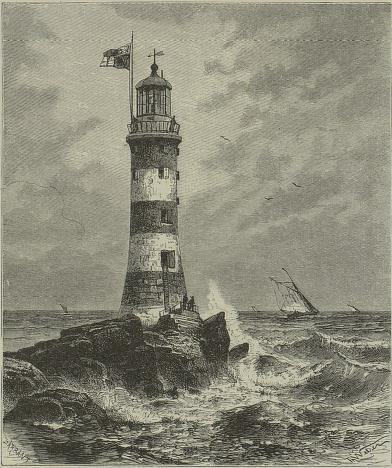} \\*
 & CLIP & \includegraphics[width=0.12\textwidth]{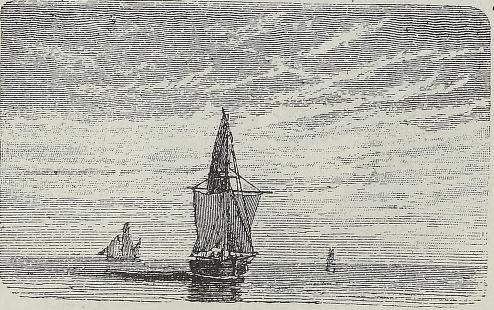} & \includegraphics[width=0.12\textwidth]{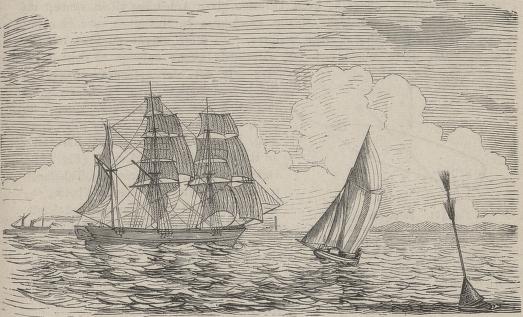} & \includegraphics[width=0.12\textwidth]{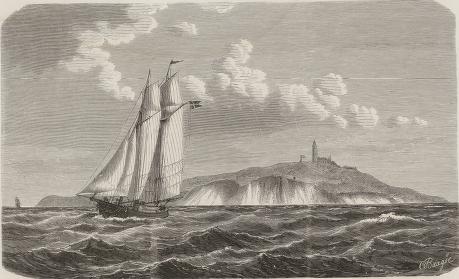} & \includegraphics[width=0.12\textwidth]{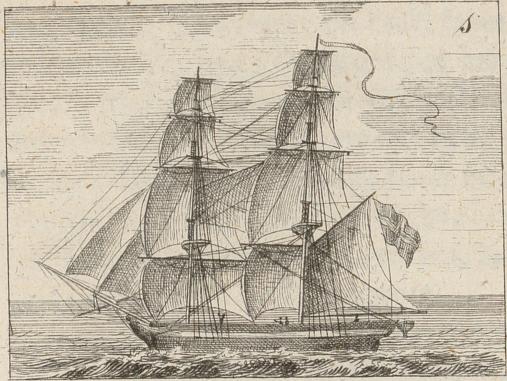} & \includegraphics[width=0.12\textwidth]{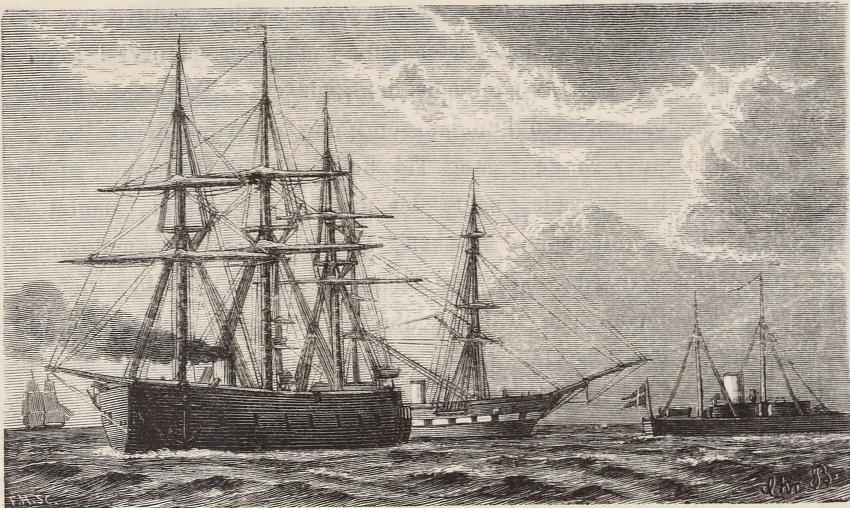} \\*
 & ViT & \includegraphics[width=0.12\textwidth]{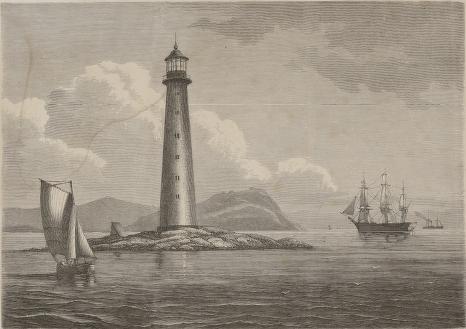} & \includegraphics[width=0.12\textwidth]{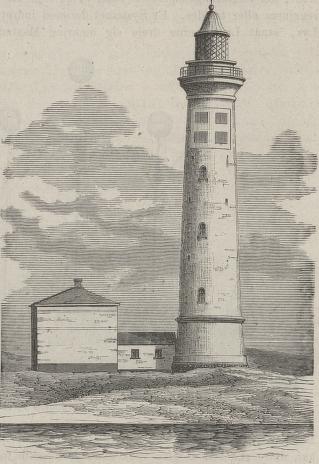} & \includegraphics[width=0.12\textwidth]{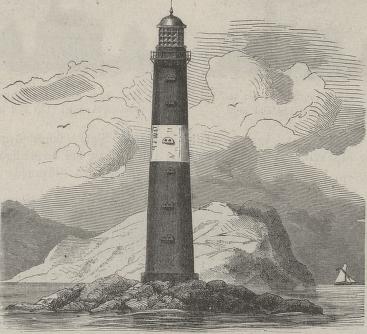} & \includegraphics[width=0.12\textwidth]{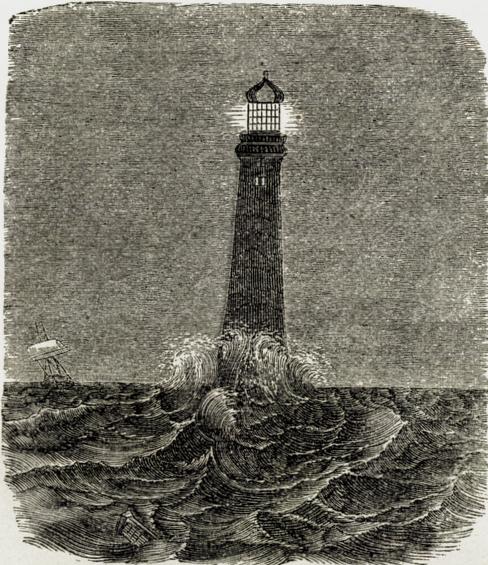} & \includegraphics[width=0.12\textwidth]{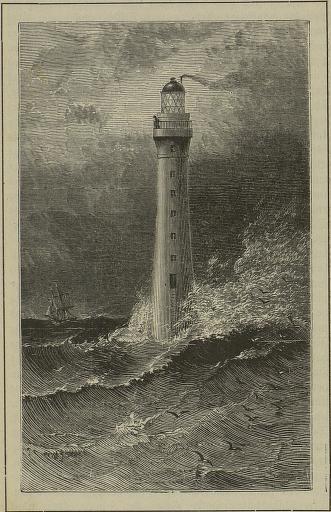} \\
\cline{1-7}
\multirow[t]{3}{*}{\includegraphics[width=0.12\textwidth]{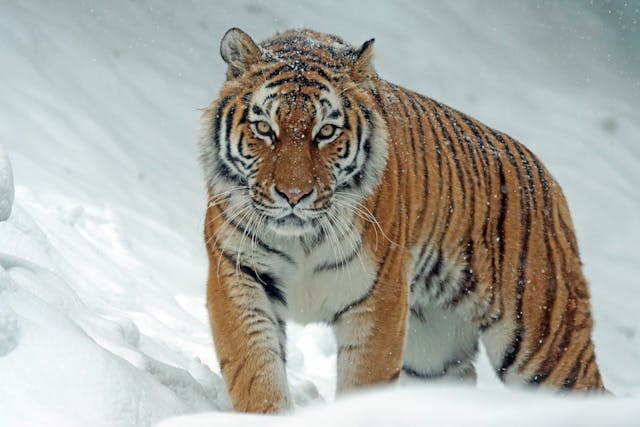}} & SigLIP & \includegraphics[width=0.12\textwidth]{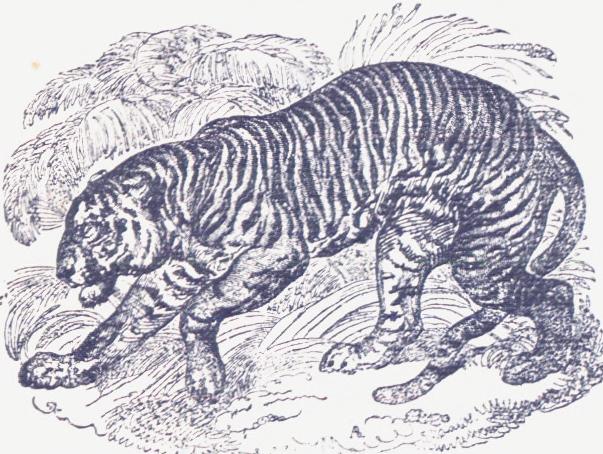} & \includegraphics[width=0.12\textwidth]{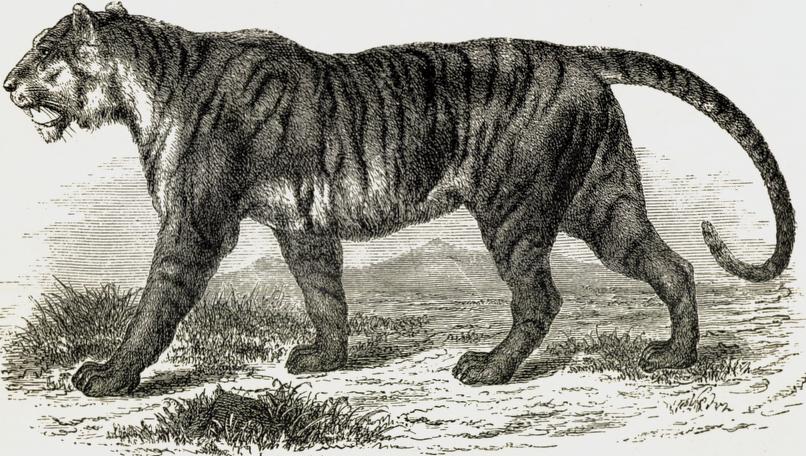} & \includegraphics[width=0.12\textwidth]{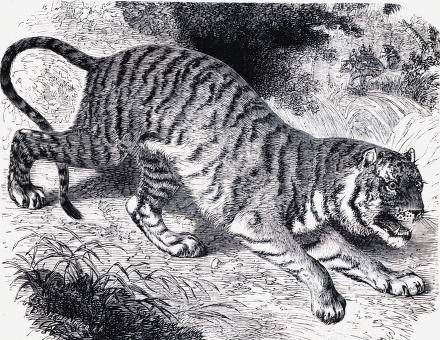} & \includegraphics[width=0.12\textwidth]{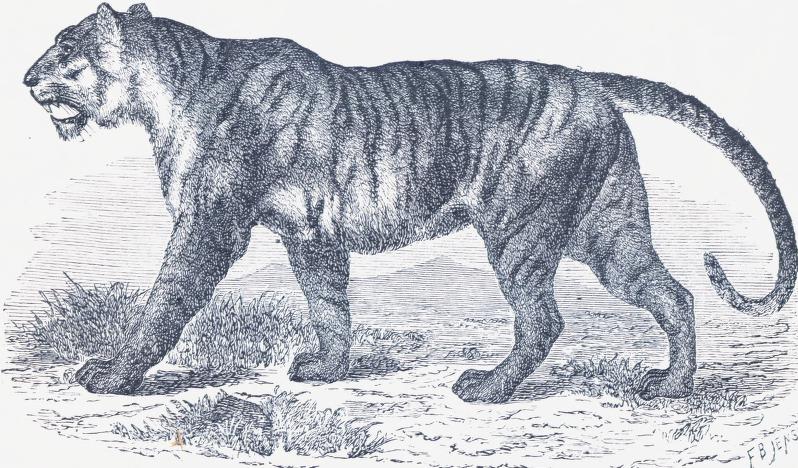} & \includegraphics[width=0.12\textwidth]{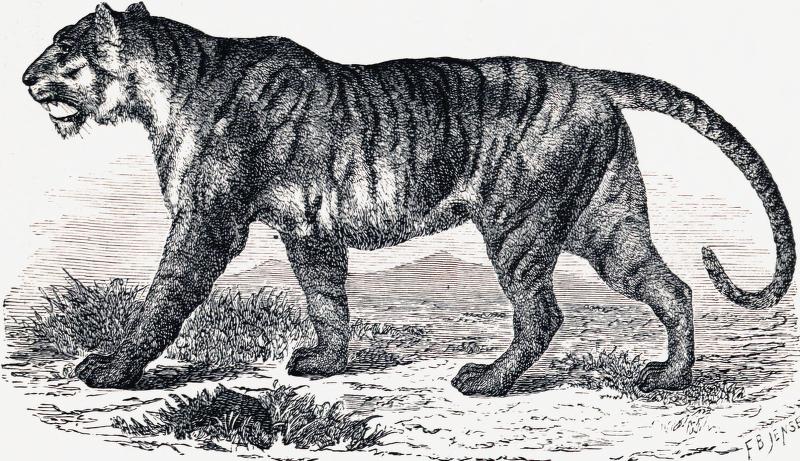} \\*
 & CLIP & \includegraphics[width=0.12\textwidth]{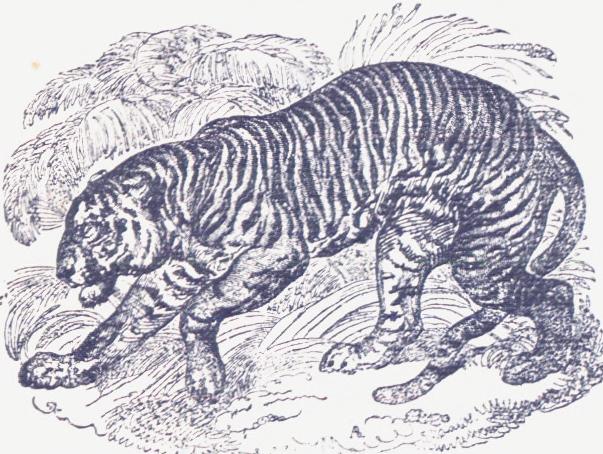} & \includegraphics[width=0.12\textwidth]{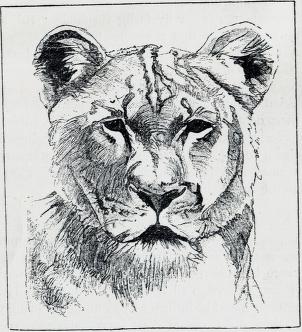} & \includegraphics[width=0.12\textwidth]{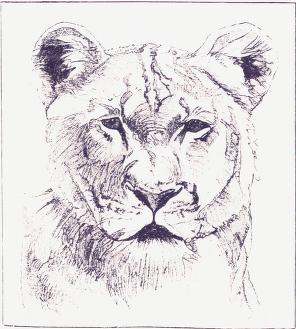} & \includegraphics[width=0.12\textwidth]{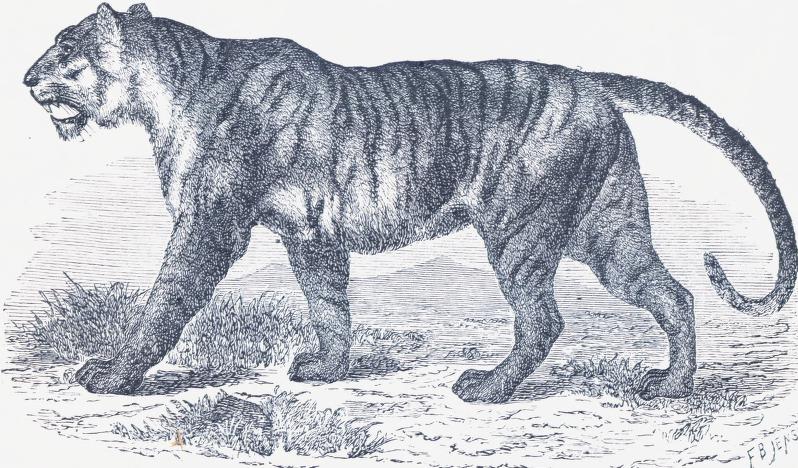} & \includegraphics[width=0.12\textwidth]{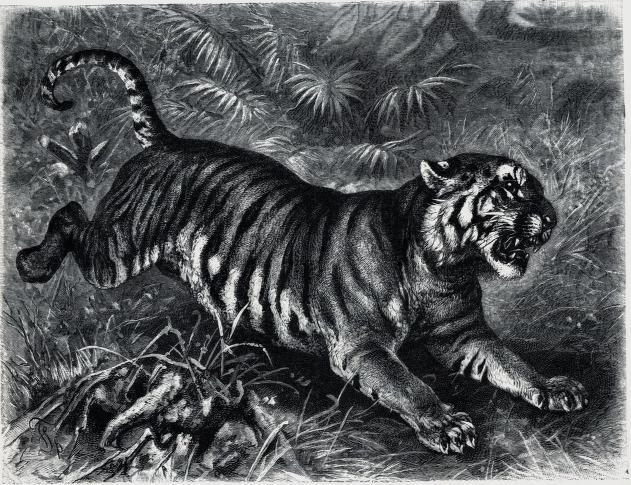} \\*
 & ViT & \includegraphics[width=0.12\textwidth]{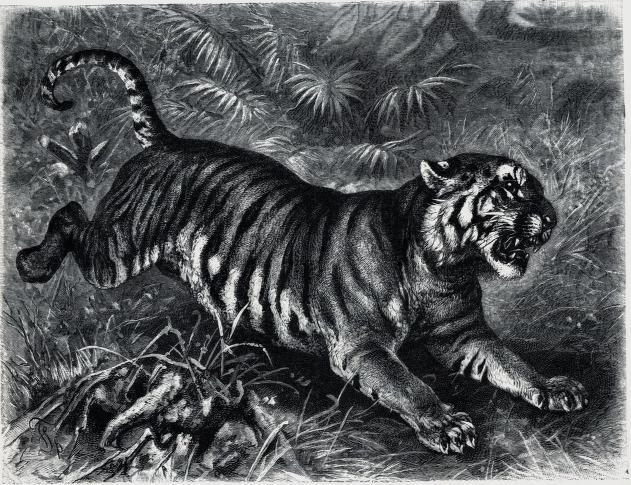} & \includegraphics[width=0.12\textwidth]{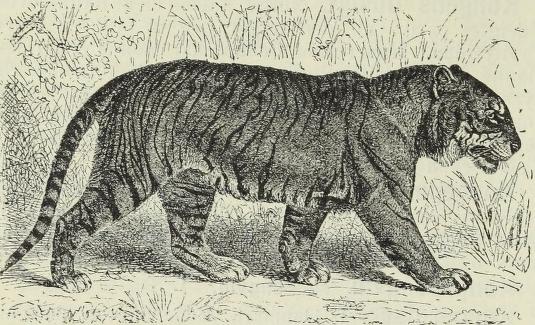} & \includegraphics[width=0.12\textwidth]{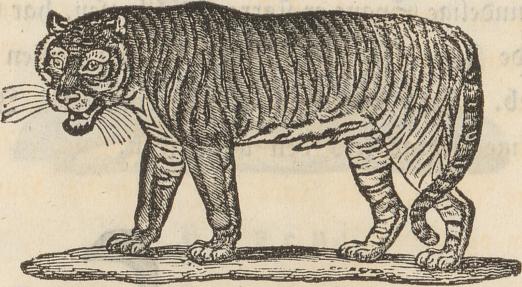} & \includegraphics[width=0.12\textwidth]{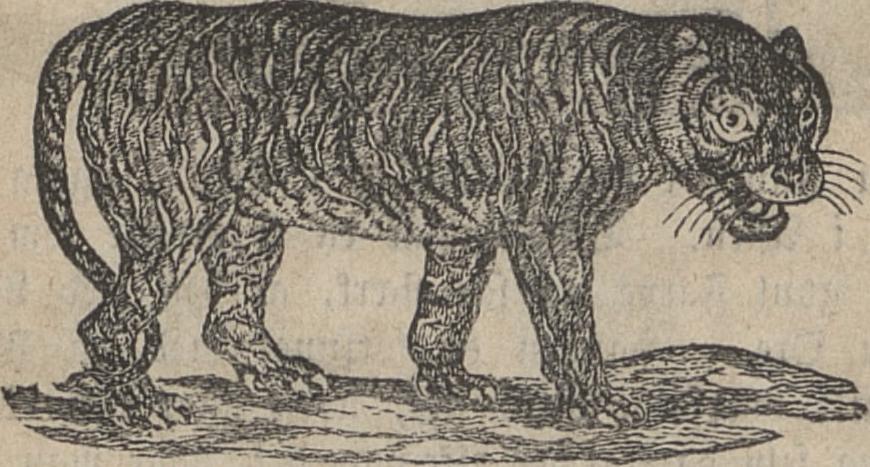} & \includegraphics[width=0.12\textwidth]{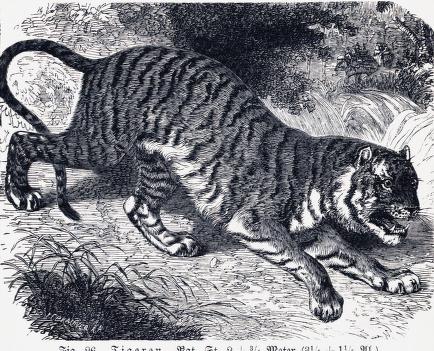} \\*
\hline \\[0ex]
\multirow[t]{3}{*}{\includegraphics[width=0.12\textwidth]{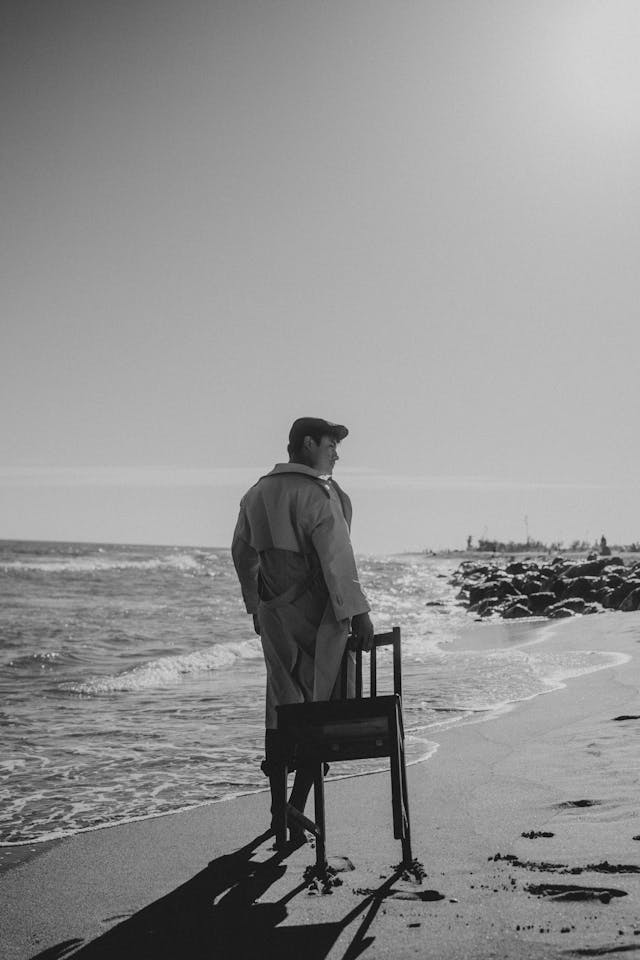}} & SigLIP & \includegraphics[width=0.12\textwidth]{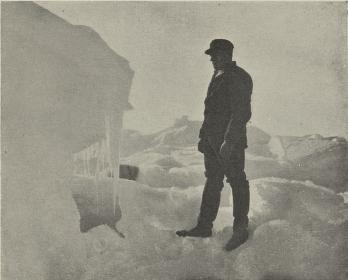} & \includegraphics[width=0.12\textwidth]{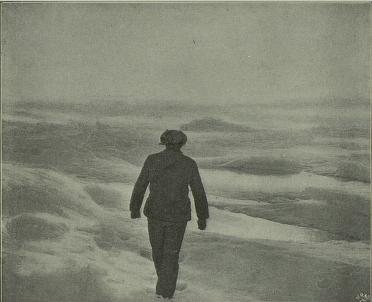} & \includegraphics[width=0.12\textwidth]{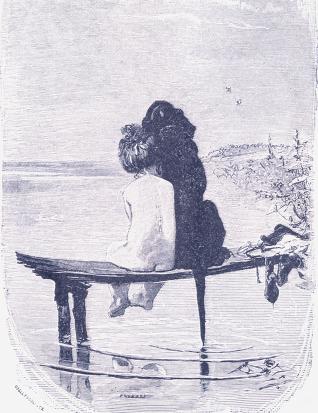} & \includegraphics[width=0.12\textwidth]{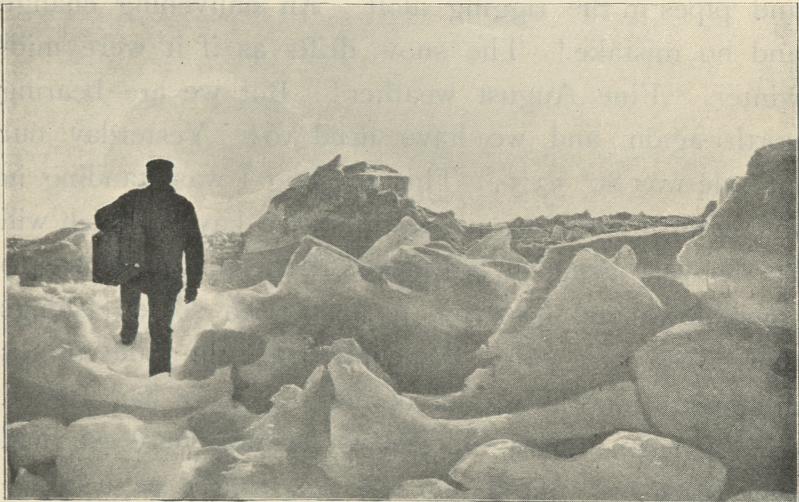} & \includegraphics[width=0.12\textwidth]{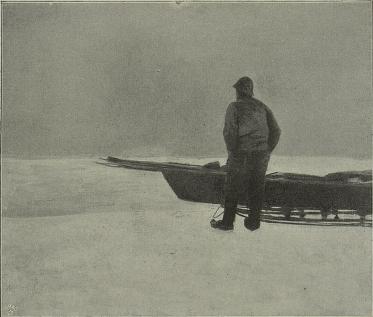} \\*
 & CLIP & \includegraphics[width=0.12\textwidth]{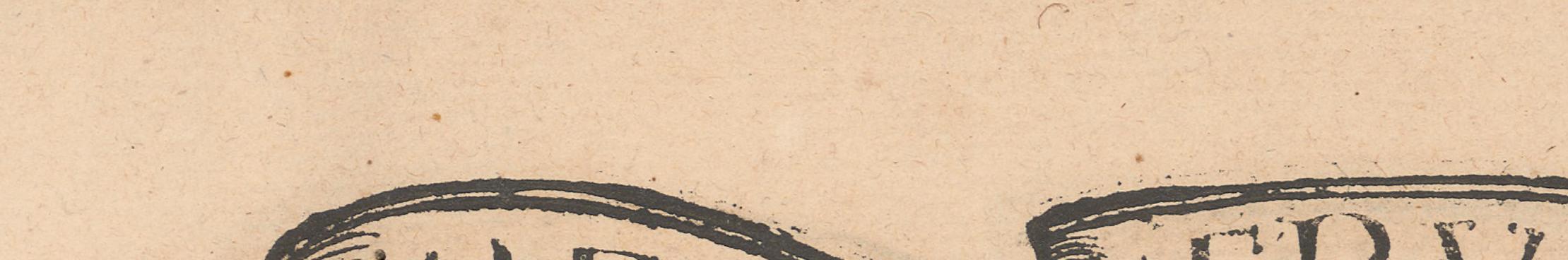} & \includegraphics[width=0.12\textwidth]{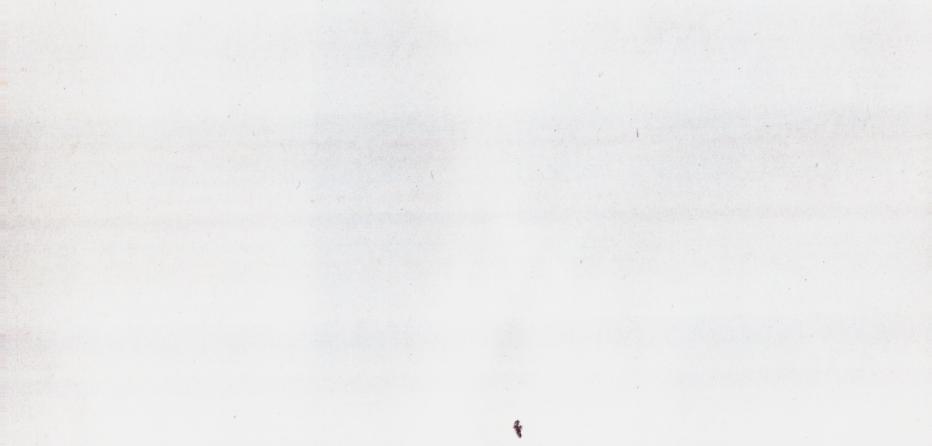} & \includegraphics[width=0.12\textwidth]{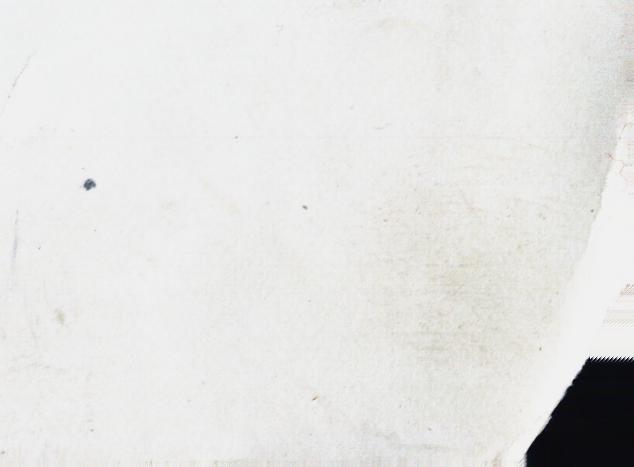} & \includegraphics[width=0.12\textwidth]{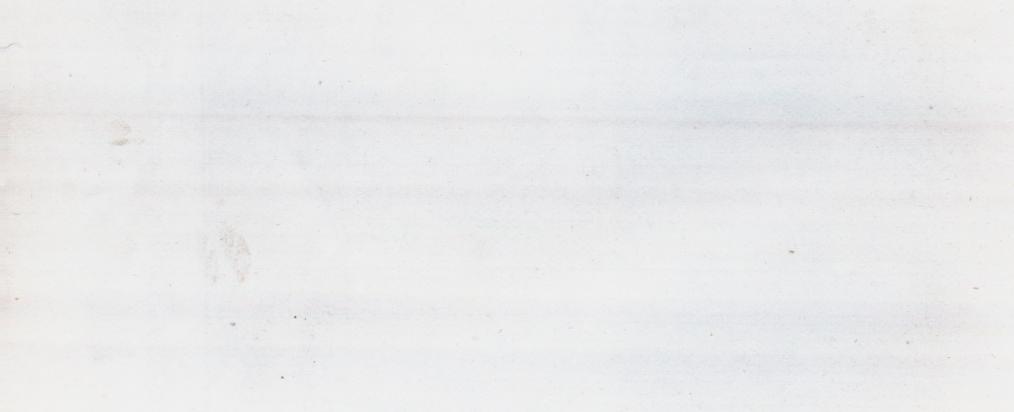} & \includegraphics[width=0.12\textwidth]{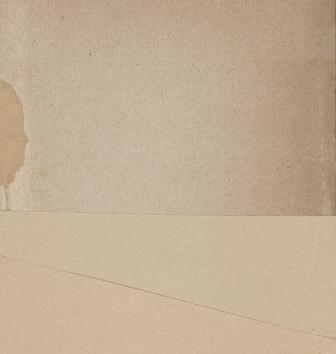} \\*
 & ViT & \includegraphics[width=0.12\textwidth]{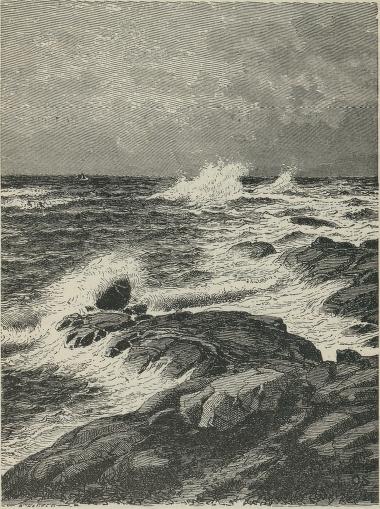} & \includegraphics[width=0.12\textwidth]{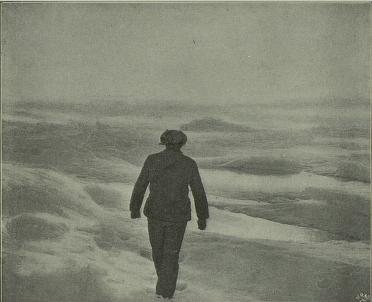} & \includegraphics[width=0.12\textwidth]{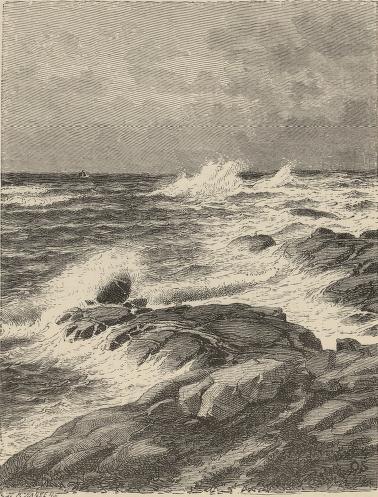} & \includegraphics[width=0.12\textwidth]{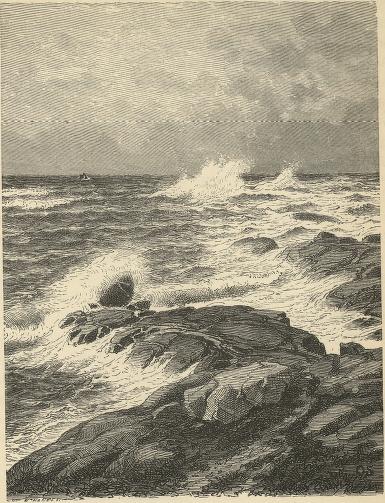} & \includegraphics[width=0.12\textwidth]{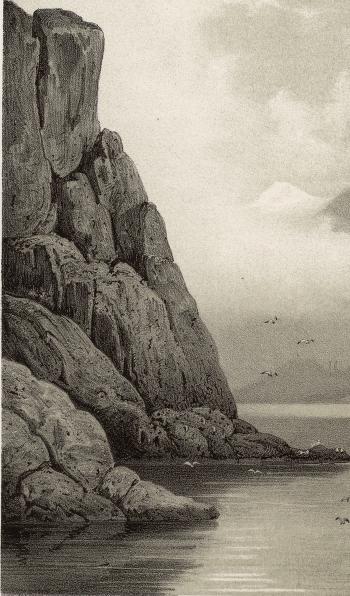} \\*
\cline{1-7}
\end{longtable}
}

\begin{table}[b]
    \centering
    \caption{Exact image retrieval accuracy}
    \label{tab:reverse.image.search}
    \begin{tabular}{lrrrr}
        \toprule
        Accuracy & Top 1 & Top 5 & Top 10 & Top 50 \\
        Model &  &  &  &  \\
        \midrule
        CLIP & \(492 \ \ (72 \ \%)\) & \(596 \ \ (87 \ \%)\) & \(613 \ \ (90 \ \%)\) & \(638 \ \ (93 \ \%)\) \\
        SigLIP & \(\mathbf{529 \ \ (77 \ \%)}\) & \(\mathbf{633 \ \ (93 \ \%)}\) & \(\mathbf{645 \ \ (94 \ \%)}\) & \(\mathbf{665 \ \ (97 \ \%)}\) \\
        ViT & \(\mathbf{529 \ \ (77 \ \%)}\) & \(582 \ \ (85 \ \%)\) & \(597 \ \ (87 \ \%)\) & \(612 \ \ (89 \ \%)\) \\
        \bottomrule
    \end{tabular}
\end{table}

\begin{table}[t]
    \centering
    \caption{Confusion matrix for the classification based on the outer cross-validation loop validation sets; it shows the number of elements with label \(a\) (columns) classified as label \(b\) (rows).}
        \begin{tabular}{@{}lrrrrrrr}
        \toprule
        \rotatebox{90}{\makecell{True\\class}} & \rotatebox{90}{\makecell{Segmentation\\anomaly}} & \rotatebox{90}{\makecell{Blank\\page}} & \rotatebox{90}{\makecell{Graphical\\element}} & \rotatebox{90}{\makecell{Illustration or\\photograph}} & \rotatebox{90}{\makecell{Musical\\notation}} & \rotatebox{90}{\makecell{Map}} & \rotatebox{90}{\makecell{Mathematical\\chart}} \\
        Predicted class &  &  &  &  &  &  &  \\
        \midrule
        Segmentation anomaly & {\cellcolor[HTML]{00682A}} \color[HTML]{F1F1F1} 496 & {\cellcolor[HTML]{F6FCF4}} \color[HTML]{000000} 5 & {\cellcolor[HTML]{F0F9ED}} \color[HTML]{000000} 28 & {\cellcolor[HTML]{F5FBF3}} \color[HTML]{000000} 8 & {\cellcolor[HTML]{F7FCF5}} \color[HTML]{000000} 2 & {\cellcolor[HTML]{F7FCF5}} \color[HTML]{000000} 1 & {\cellcolor[HTML]{F7FCF5}} \color[HTML]{000000} 2 \\
        Blank page & {\cellcolor[HTML]{F4FBF2}} \color[HTML]{000000} 11 & {\cellcolor[HTML]{48AE60}} \color[HTML]{F1F1F1} 339 & {\cellcolor[HTML]{F5FBF3}} \color[HTML]{000000} 8 & {\cellcolor[HTML]{F7FCF5}} \color[HTML]{000000} 1 & {\cellcolor[HTML]{F7FCF5}} \color[HTML]{000000} 0 & {\cellcolor[HTML]{F7FCF5}} \color[HTML]{000000} 0 & {\cellcolor[HTML]{F7FCF5}} \color[HTML]{000000} 0 \\
        Graphical element & {\cellcolor[HTML]{F4FBF1}} \color[HTML]{000000} 14 & {\cellcolor[HTML]{F7FCF5}} \color[HTML]{000000} 2 & {\cellcolor[HTML]{75C477}} \color[HTML]{000000} 278 & {\cellcolor[HTML]{F4FBF1}} \color[HTML]{000000} 15 & {\cellcolor[HTML]{F7FCF5}} \color[HTML]{000000} 1 & {\cellcolor[HTML]{F7FCF5}} \color[HTML]{000000} 0 & {\cellcolor[HTML]{F7FCF5}} \color[HTML]{000000} 2 \\
        Illustration or photograph & {\cellcolor[HTML]{F7FCF5}} \color[HTML]{000000} 1 & {\cellcolor[HTML]{F6FCF4}} \color[HTML]{000000} 3 & {\cellcolor[HTML]{F3FAF0}} \color[HTML]{000000} 16 & {\cellcolor[HTML]{00441B}} \color[HTML]{F1F1F1} 558 & {\cellcolor[HTML]{F7FCF5}} \color[HTML]{000000} 1 & {\cellcolor[HTML]{F7FCF5}} \color[HTML]{000000} 2 & {\cellcolor[HTML]{F6FCF4}} \color[HTML]{000000} 5 \\
        Musical notation & {\cellcolor[HTML]{F7FCF5}} \color[HTML]{000000} 1 & {\cellcolor[HTML]{F7FCF5}} \color[HTML]{000000} 0 & {\cellcolor[HTML]{F7FCF5}} \color[HTML]{000000} 0 & {\cellcolor[HTML]{F7FCF5}} \color[HTML]{000000} 0 & {\cellcolor[HTML]{D4EECE}} \color[HTML]{000000} 109 & {\cellcolor[HTML]{F7FCF5}} \color[HTML]{000000} 0 & {\cellcolor[HTML]{F7FCF5}} \color[HTML]{000000} 0 \\
        Map & {\cellcolor[HTML]{F7FCF5}} \color[HTML]{000000} 1 & {\cellcolor[HTML]{F7FCF5}} \color[HTML]{000000} 0 & {\cellcolor[HTML]{F7FCF5}} \color[HTML]{000000} 0 & {\cellcolor[HTML]{F7FCF5}} \color[HTML]{000000} 2 & {\cellcolor[HTML]{F7FCF5}} \color[HTML]{000000} 0 & {\cellcolor[HTML]{EDF8E9}} \color[HTML]{000000} 41 & {\cellcolor[HTML]{F7FCF5}} \color[HTML]{000000} 0 \\
        Mathematical chart & {\cellcolor[HTML]{F7FCF5}} \color[HTML]{000000} 0 & {\cellcolor[HTML]{F7FCF5}} \color[HTML]{000000} 0 & {\cellcolor[HTML]{F7FCF5}} \color[HTML]{000000} 0 & {\cellcolor[HTML]{F5FBF3}} \color[HTML]{000000} 8 & {\cellcolor[HTML]{F7FCF5}} \color[HTML]{000000} 0 & {\cellcolor[HTML]{F7FCF5}} \color[HTML]{000000} 0 & {\cellcolor[HTML]{EDF8EA}} \color[HTML]{000000} 39 \\
        \bottomrule
        \end{tabular}
        
        \begin{tablenotes}
            \item A perfect classifier will only have nonzero entries on the diagonal.
        \end{tablenotes}
    \label{tab:confusion.matrix}
\end{table}
\appendix

\end{document}